\pdfoutput=1

\documentclass[11pt]{article}

\usepackage[]{acl2023}

\usepackage{times}
\usepackage{latexsym}
\usepackage{comment}
\usepackage{booktabs}

\usepackage[hang,flushmargin]{footmisc} %

\usepackage[T1]{fontenc}
\usepackage[utf8]{inputenc}

\usepackage{microtype}

\usepackage{inconsolata}

\title{AlbNER: A Corpus for Named Entity Recognition in Albanian}

\author{Erion \c{C}ano \\
  Digital Philology \\
  Data Mining and Machine Learning \\
  University of Vienna, Austria \\
  \texttt{erion.cano@univie.ac.at} \\
  }

\begin{document}

\maketitle

\begin{abstract}
Scarcity of resources such as annotated text corpora for under-resourced languages like Albanian is a serious impediment in computational linguistics and natural language processing research. This paper presents AlbNER, a corpus of 900 sentences with labeled named entities, collected from Albanian Wikipedia articles. Preliminary results with BERT and RoBERTa variants fine-tuned and tested with AlbNER data indicate that model size has slight impact on NER performance, whereas language transfer has a significant one. AlbNER corpus and these obtained results should serve as baselines for future experiments.   
\end{abstract}

\section{Introduction}

Data-driven artificial intelligence methods have advanced rapidly during the last decade. Two directions that have especially witnessed stunning progress are the ones related to image and text or natural language processing. Trained artificial neural networks are providing excellent results when solving tasks like image segmentation, object detection, medical image analysis and more from the former, and machine translation, text summarization, sentiment analysis and more from the later. There are even some tasks that put images and text together such as image captioning \cite{sharma-etal-2018-conceptual} or meme processing \cite{10.1007/978-3-031-41676-7_31} which have driven significant innovation. 

Within NLP (Natural Language Processing), machine translation has been the task that has driven the biggest leaps, especially since the introduction of the Transformer architecture \cite{NIPS2017_3f5ee243} and the PLMs (Pretrained Language Models) such as BERT \cite{devlin-etal-2019-bert}. PLMs got quick adaptation and have become the defacto standard paradigm for solving many other NLP tasks, including NER (Named Entity Recognition) which tries to identify and classify key information words or phrases in texts. 

The main strategy of preparing and utilizing PLMs is by pretraining Transformer blocks (layers) with large amounts of unlabeled texts (not related to any specific task), and later fine-tuning them with labeled texts which are suitable for solving a specific task. In the case of NER, the labeled texts comprise tokens and labels such as PER (person), LOC (location), ORG (organization) etc., which identify the category of each token \cite{Jain_2020}. 

One difficulty when working with language models is the need for large amount of pretraining texts and the need for labeled and task-specific corpora which are used during the fine-tuning phase. While not being a problem for languages like English, Spanish, Chinese, etc., the latter is a significant restriction when trying to solve tasks for other ``smaller'' languages, which are also known as low-resource or underrepresented \cite{aji-etal-2022-one}. Unavailability of such resources limits the progress and the attainable performance on natural language processing and computational linguistics tasks for those languages.  

This paper presents AlbNER, a corpus of sentences in Albanian language, created with the goal to foster research in named entity recognition.\footnote{Download from: \url{http://hdl.handle.net/11234/1-5214}} They were collected from Wikipedia and cover topics related to Albanian history and geography, as well as Albanian historic figures. There are also generic sentences about common facts. Each sentence was tokenized and the resulting tokens were labeled manually. In total, a set of 900 samples (sentences) was obtained.   

A set of experiments were run, assessing the performance of a few PLMs fine-tuned and tested with AlbNER data. The results are overall poor, since the PLMs have not been pretrained with Albanian texts. The best results were actually obtained using a multilingual model which was pretrained on many languages, including Albanian. This indicates that knowledge transfer has a high impact on NER performance in Albanian. Another insight is the fact that model size has low impact, since the score differences between the large and the base versions of the tested PLMs were small. These obtained results should serve as baselines for further research in NER for Albanian.

\section{Related Work} %

NER has been a popular task for several years, since it helps to extract relevant and important information. It usually works as a two-step process, starting with the detection of the named entities (words or phrases) in the text, and then classifying them in predefined categories such as \textit{person}, \textit{location}, \textit{organization}, etc. 

Some of the earliest solutions that were proposed came out in the nineties and were based on supervised learning techniques such as Decision Trees \cite{sekine-etal-1998-decision}, Hidden Markov Models \cite{bikel-etal-1997-nymble}, Maximum Entropy Models \cite{borthwick-etal-1998-exploiting}, etc. Here, relatively small amounts of sentences were manually labeled and used for training the popular supervised learning algorithms of that time.

\begin{table}[t]
\centering
\begin{tabular}{l c c}
\hline
\textbf{Data} & \textbf{Sentences} & \textbf{Tokens} \\
\hline
Train & 500 & 8826 \\
Dev & 100 & 1732 \\
Test & 300 & 5266 \\
Total & 900 & 15824 \\ \hline
\end{tabular}
\caption{Sentences and tokens in AlbNER.}
\label{tab:stats}
\end{table}

Latter on, several attempts were made by integrating knowledge from lexicons \cite{alfonseca-manandhar-2002-proposal,passos-etal-2014-lexicon}, by trying semi-supervised methods \cite{althobaiti-etal-2015-combining} or probing unsupervised learning techniques \cite{bhagavatula-etal-2012-language,suzuki-etal-2011-learning}. Utilizing knowledge bases with Wikipedia resources was also explored \cite{richman-schone-2008-mining}. The most recent works obviously use PLMs such as BERT \cite{SUN2021103799}, or combinations of PLMs and other structures \cite{LIU2021106958}.

With respect to the targe language, the NER methods are monolingual, bilingual or multilingual. Most of the works involve English and other popular natural languages. As for the under-represented languages, there has been some attempts to create corpora in Hungarian \cite{10.1007/978-3-030-83527-9_19}, Upper Sorbian \cite{howson_2017}, Kashubian \cite{Nomachi+2019+453+490}, Czech \cite{DBLP:journals/corr/abs-2105-11314,nlpinai19} etc. Some works have also created resources in Albanian, but they are limited to other tasks such as sentiment analysis \cite{DBLP:journals/corr/abs-2306-08526}.

\begin{table}[t]
\centering
\begin{tabular}{l c c c c c}
\hline
\textbf{Data} & \textbf{PER} & \textbf{LOC} & \textbf{ORG} & \textbf{MISC} & \textbf{all} \\
\hline
Train & 452 & 486 & 321 & 445& 1704 \\
Dev & 92 & 73 & 68 & 78 & 311 \\
Test & 246 & 264 & 141 & 300 & 951 \\
Total & 790 & 823 & 530 & 823 & 2966 \\ \hline
\end{tabular}
\caption{Counts of named entity labels in AlbNER.}
\label{tab:necounts}
\end{table}

\begin{table}[t]
\centering
\begin{tabular}{l c c c c c}
\hline
\textbf{Data} & \textbf{PER} & \textbf{LOC} & \textbf{ORG} & \textbf{MISC} & \textbf{all}  \\
\hline
Train & 0.051 & 0.055 & 0.036 & 0.05 & 0.193 \\
Dev & 0.053 & 0042 & 0.039 & 0.045 & 0.179 \\
Test & 0.046 & 0.05 & 0.026 & 0.056 & 0.18 \\
Total & 0.05 & 0.052 & 0.033 & 0.052 & 0.187 \\ \hline
\end{tabular}
\caption{Densities of named entity labels in AlbNER.} %
\label{tab:nedensities}
\end{table}

\section{AlbNER Corpus} %

Wikipedia pages usually contain information about history and historical figures (persons), geography and places (locations), as well as other named entities. They therefore represent good sources for building NER copora. To build AlbNER, pages from Albanian Wikipedia dump were used.\footnote{\url{https://dumps.wikimedia.org/sqwiki/latest/}} A total of 900 sentences were extracted and manually curated, fixing typos and other issues. 

The sentences were later tokenized and a manual process of NER tagging followed. For the latter, CoNLL-2003 shared task annotation scheme was used.\footnote{\url{https://www.cnts.ua.ac.be/conll2003/ner/annotation.txt}} It mandates the use of O for non-entity tokens, as well as B-PER and I-PER, B-LOC and I-LOC, B-ORG and I-ORG, and finally B-MISC and I-MISC for persons, locations, organizations and other types of named entities. The initial B and I symbols show that the token is either the beggining of a named entity phrase, or appears inside of it. 

\begin{table*}[t]
\begingroup
\setlength{\tabcolsep}{9pt} %
\centering
\begin{tabular}{| l l }
\hline
\textbf{Token} & \textbf{Tag} \\
\hline
U & O \\
lind & O \\ 
më & O \\
11 & O \\
prill & O \\
1872 & O \\
në & O \\
Drenovë & B-LOC \\
, & O \\
5 & O \\
km & O \\
larg & O \\
Korçës & B-LOC \\
. & O \\ \hline
\end{tabular}
\begin{tabular}{ l l }
\hline
\textbf{Token} & \textbf{Tag} \\
\hline
Bregdeti & B-LOC \\
i & I-LOC \\
Adriatikut & I-LOC \\
shtrihet & O \\
nga & O \\
Gryka & B-LOC \\
e & I-LOC \\
Bunës & I-LOC \\
deri & O \\
në & O \\
Kepin & B-LOC \\
e & I-LOC \\
Gjuhëzës & I-LOC \\
. & O \\ \hline
\end{tabular}
\begin{tabular}{ l l }
\hline
\textbf{Token} & \textbf{Tag} \\
\hline
U & O \\
shqua & O \\
si & O \\
prijës & O \\
ushtarak & O \\
me & O \\
kontribut & O \\
në & O \\
mbrojtjen & O \\
e & O \\
Plavës & B-LOC \\
dhe & O \\
Gucisë & B-LOC \\
. & O \\ \hline
\end{tabular}
\begin{tabular}{ l l |}
\hline
\textbf{Token} & \textbf{Tag} \\
\hline
Më & O \\
1451 & O \\
, & O \\
u & O \\ 
martua & O \\
me & O \\
Donikën & B-PER \\
, & O \\
të & O \\
bijën & O \\
e & O \\
Gjergj & B-PER \\
Arianitit & I-PER \\
. & O \\ \hline
\end{tabular}
\caption{Illustration of four AlbNER sentences with their respective tokens and the NER tags.}
\label{tab:examples}
\endgroup
\end{table*}

The obtained 900 samples were divided in three subsets for model training, developmen and testing. Their respective token counts are shown in Table~\ref{tab:stats}. The average sentence length is 17.58 tokens. Furthermore, Table~\ref{tab:necounts} shows the token count distribution per each named entity category, in each of the AlbNER cuts. The respective named entity densities are summarized in Table~\ref{tab:nedensities}. Finally, Table~\ref{tab:examples} illustrates four samples of AlbNER corpus.

\section{Preliminary Experimental Results} \label{sec:results}

This section presents the results of some basic experiments that were run using AlbNER corpus and a few PLMs fine-tuned for named entity recognition. These results are intended as baselines for future studies. 

\subsection{Experimental Setup} \label{ssec:setup}

\begin{table}[t]
\centering
\begin{tabular}{l c}
\hline
\textbf{Hyperparemeter} & \textbf{Values} \\
\hline
epochs small ($E_s$) & 5 \\
epochs total ($E_t$) & 10 \\
top models ($T$) & 3 \\
batch size & 16 \\
gradient accumulation steps & [4, 8] \\
learning rate & [1e-4, 1e-5] \\
cfr & [True, False] \\
weitht decay & 1e-7 \\
warmup step ratio & 0.1 \\
max gradient norm & 10 \\ \hline
\end{tabular}
\caption{Searched values for each hyperparameter.}
\label{tab:hyperparams}
\end{table} 

To easily experiment with PLMs on the NER task, T-NER framework was utilized \cite{ushio-camacho-collados-2021-ner}. It provides software utilities to fine-tune and evaluate various models based on popular PLMs using custom data. 
The tested models were based on BERT and RoBERTa \cite{DBLP-roberta} which are two very popular PLMs of the recent years. Two versions of the former were used: BERT-base with 110\,M parameters and BERT-large with 340\,M parameters. RoBERTa differs from BERT in several training specifications and in the fact that it was pretrained on bigger texts. Its two main versions, RoBERTa-base and RoBERTa-large have 123\,M and 354\,M parameters respectively. Probing both base and large versions of these PLMs gives us an idea of how much does the NER task benefit from model size. 

It is also important to note that all these four PLMs were pretrained on English texts and fine-tuned on AlbNER texts which are in Albanian. To have an idea about the knowledge transfer between the two languages and the possible benefits from multilingual pretraining, another version of BERT (denoted BERT-base-ML) was used. This last PLM has been pretrained on Wiki texts of 104 languages, one of which is Albanian. 

Each of the five PLMs was fine-tunned on AlbNER following a parameter-search procedure which consists of two steps: (1) fine-tunning with every possible configuration $C$ (obtained from the combination of the searched hyperparameters) for a small number of epochs $E_s$ and computing micro $F_1$ metric on the development set, and (ii) picking the top $T$ models to continue fine-tunning till $E_t$ epochs. The best model in the second stage will continue fine-tunning as long as it improves on the development set. The list of fine-tunning hyperparameters are shown in Table~\ref{tab:hyperparams}. Searching more values for each hyperparameter could lead to better results, but it would take significantly more time.

\begin{table*}[t]
\centering
\begin{tabular}{l c c c c c}
\hline
\textbf{Model} & \textbf{Pretrain} & \textbf{Fine-tune} & \textbf{micro F$_1$} & \textbf{macro F$_1$} & \textbf{weighted F$_1$} \\
\hline
BERT-base & English & Albanian & 0.45 & 0.42 & 0.44 \\
BERT-large & English & Albanian & 0.49 & 0.46 & 0.48 \\
RoBERTa-base & English & Albanian & 0.52 & 0.41 & 0.48 \\
RoBERTa-large & English & Albanian & 0.57 & 0.53 & 0.57 \\
BERT-base-ML & Multiple & Albanian & \textbf{0.62} & \textbf{0.59} & \textbf{0.61} \\ \hline
\end{tabular}
\caption{NER results of various pretrained language models tuned and tested with AlbNER.}
\label{tab:results}
\end{table*}

\subsection{Discussion of Results} \label{ssec:discussion}

Each of the five models described above was tested on the 300 samples of the test cut of AlbNER. The micro $F_1$, macro $F_1$ and the weighted $F_1$ scores obtained for each of them are shown in Table~\ref{tab:results}. 

As can be seen, the results are overall poor, given that $F_1$ levels of 0.9 or higher can be achived if we fine-tune and test on English corpora \cite{luoma-pyysalo-2020-exploring}. Knowledge transfer between English and Albanian is obviously limited. We see that BERT-base which is the smallest performs worse than the other models, with a weighted $F_1$ of 0.44. BERT-large slightly improves over it. RoBERTa-base is smaller than BERT-large in terms of parameters, but has been pretrained in larger text corpora. It slightly outruns BERT-large. 

RoBERTa-large improves fairly over RoBERTa-base, reaching a weighted $F_1$ of 0.57. The best model is BERT-base-ML with weighted $F_1$ of 0.61. Despite being small in number of parameters, it has been pretrained in many languages, including Albanian. This clearly shows that knowledge transfer has a significant impact on NER performance. For this reason, better NER results could be achived by utilizing PLMs which are completely pretrained with Albanian texts, which represents possible future extension of this work.

\section{Conclusions} \label{sec:conclusions}

Because of the recent data-driven trends for solving natural language processing tasks, resources such as curated and annotated text corpora have become indispensible. To foster research on named entity recognition in Albanian language, this work creates and presents AlbNER, a corpus of 900 sentences collected from Wikipedia articles which were labeled following CoNLL-2003 annotation scheme. A set of experiments were conducted, utilizing BERT and RoBERTa variants. The preliminary results indicate that NER performance is overall poor if PLMs pretrained with English texts are used. They also indicate that the task benefits little from PLM size, but significantly from language transfer during pretraining and fine-tuning.  

\bibliography{custom}
\bibliographystyle{acl_natbib}

\end{document}